\documentclass[sigconf]{aamas}

\usepackage{balance} 

\usepackage{latexsym}
\usepackage{amsmath}
\usepackage{amsthm}
\usepackage{booktabs}
\usepackage{enumitem}
\usepackage{graphicx}
\usepackage{color}
\usepackage[acronym]{glossaries}
\usepackage{xfrac}
\usepackage[caption=false]{subfig}

\usepackage{siunitx}
\usepackage{mathtools}
\usepackage{bbold}
\usepackage{url}
\usepackage{bm}
\usepackage{multicol}
\usepackage{incgraph,tikz}
\usepackage{calrsfs}   
\usepackage{algorithm}
\usepackage[noend]{algpseudocode}
\DeclareMathAlphabet{\pazocal}{OMS}{zplm}{m}{n}

\newcommand{\BibTeX}{B\kern-.05em{\sc i\kern-.025em b}\kern-.08em\TeX}

\newcommand{\N}{\mathbb N}
\DeclareMathOperator*{\argmax}{argmax}

\setcopyright{ifaamas}
\acmConference[AAMAS '25]{Proc.\@ of the 24th International Conference
on Autonomous Agents and Multiagent Systems (AAMAS 2025)}{May 19 -- 23, 2025}
{Detroit, Michigan, USA}{A.~El~Fallah~Seghrouchni, Y.~Vorobeychik, S.~Das, A.~Nowe (eds.)}
\copyrightyear{2025}
\acmYear{2025}
\acmDOI{}
\acmPrice{}
\acmISBN{}

\acmSubmissionID{738}

\title[VO-MCTS planning in dynamic environments]{Monte Carlo Tree Search with Velocity Obstacles for safe and efficient motion planning in dynamic environments}

\author{Lorenzo Bonanni$^{\star}$}
\affiliation{
  \institution{University of Verona}
  \city{Verona}
  \country{Italy}}
\email{lorenzo.bonanni@univr.it}

\author{Daniele Meli$^{\star}$}
\affiliation{
  \institution{University of Verona}
  \city{Verona}
  \country{Italy}}
\email{daniele.meli@univr.it}

\author{Alberto Castellini}
\affiliation{
  \institution{University of Verona}
  \city{Verona}
  \country{Italy}}
\email{alberto.castellini@univr.it}

\author{Alessandro Farinelli}
\affiliation{
  \institution{University of Verona}
  \city{Verona}
  \country{Italy}}
\email{alessandro.farinelli@univr.it}

\begin{abstract}
Online motion planning is a challenging problem for intelligent robots moving in dense environments with dynamic obstacles, e.g., crowds.
In this work, we propose a novel approach for optimal and safe online motion planning with minimal information about dynamic obstacles. Specifically, our approach requires only the current position of the obstacles and their maximum speed, but it does not need any information about their exact trajectories or dynamic model. 
The proposed methodology combines Monte Carlo Tree Search (MCTS), for online optimal planning via model simulations, with Velocity Obstacles (VO), for obstacle avoidance. 
We perform experiments in a cluttered simulated environment with walls, and up to 40 dynamic obstacles moving with random velocities and directions.
With an ablation study, we show the key contribution of VO in scaling up the efficiency of MCTS, selecting the safest and most rewarding actions in the tree of simulations. Moreover, we show the superiority of our methodology with respect to state-of-the-art planners, including Non-linear Model Predictive Control (NMPC), in terms of improved collision rate, computational and task performance.
\end{abstract}


\begin{document}


\pagestyle{fancy}
\fancyhead{}


\maketitle 


\section{Introduction}
Motion planning for mobile robots is an important and widely studied research area.
When robots move in a (possibly uncertain) environment in the presence of dynamic obstacles, e.g., other agents or people, they must balance trajectory optimality towards the goal and risk of collision.
Fast real-time trajectory computation is a key feature in dynamic environments to guarantee prompt response and adaptation to changes in the environment~\cite{mohanan2018survey}. 

\emph{Reactive planning} methods, e.g., Velocity Obstacles (VO) \cite{fiorini1998motion} and Artificial Potential Fields \cite{warren1989global} consider a single motion command per time step, but they can get stuck in local minima when maps are complex \cite{fan2020improved}. On the other hand, \emph{look-ahead planning methods}, as Non-linear Model Predictive Control (NMPC) \cite{liu2017path} and tree-based search \cite{salzman2016asymptotically}, are more robust since they optimize the trajectory over a time horizon. However, they are computationally demanding in dynamic environments, where re-planning is often needed. Furthermore, these methods often require precise knowledge about the trajectories of dynamic obstacles \cite{tordesillas2021mader}, which is often unavailable or uncertain in real-world domains, especially when dealing with agents exhibiting heterogeneous behaviors (e.g., humans \cite{gupta2022intention}). Deep Reinforcement Learning (DRL) has recently become popular for motion planning in complex dynamic environments; however, it often struggles to generalize beyond the training scenario, offering no guarantee about safe collision avoidance. Additionally, its effective implementation requires extensive training data \cite{amir2023verifying}.

In this paper, we propose a novel approach to online robotic motion planning in dense, dynamic and partially unknown environments, combining look-ahead and reactive planning.
Unlike other methods assuming partially known obstacle trajectories \cite{tordesillas2021mader,li2020socially}, our approach only requires knowledge about the instantaneous obstacle locations, which can be obtained from standard sensors (e.g., LIDARs and cameras), and an upper bound on their maximum velocity (typically available, e.g., from social models \cite{karamouzas2014universal} or other robots' specifications).
We define the problem as a Markov Decision Process (MDP) and solve it using Monte Carlo Tree Search (MCTS), which performs online simulations over a time horizon to estimate the long-term expected value of actions and determine the next best action.
MCTS often fails to scale to large action spaces \cite{dam2022monte,Castellini2023,Bianchi_AIRO_2022}, which limits its applicability to fine-control mobile robots' velocity.
For this reason, we combine MCTS with the VO paradigm, which prunes unsafe actions (i.e., leading to collisions) from the search space during MCTS simulations, reducing the action search space. If the positions and maximum velocities of obstacles are known\footnote{In case of uncertainties, upper bounds can be used.}, the integration of the two approaches guarantees that the agent always picks \emph{safe (i.e., not colliding) and optimal actions}, considering a sufficiently small time-step
Crucially, using the proposed approach, the number of simulations required by MCTS is significantly reduced, fostering the deployment of the approach to real robotic platforms, which often have limited computational resources.

We validate our methodology in a simulated $10 \times 10$ \si{m} map (inspired from \cite{gupta2022intention}) containing up to 40 randomly moving obstacles with a diameter of $0.2$ \si{m} each.
Our results show that VO allows to greatly improve the performance of MCTS, achieving higher cumulative reward and success rate (i.e., collision-free goal reaching) even with very few simulations (up to 10 simulations per MCTS step, corresponding to a planning time of $\approx$\SI{0.02}{s}). 
Furthermore, the proposed approach achieves superior cumulative reward with fewer collisions with respect to competitor motion planners, including the classical Dynamic Window Approach (DWA) \cite{ liu2021path} and NMPC \cite{liu2017path}, an established methodology for optimal motion planning \cite{tordesillas2021mader} which is considerably more computationally demanding.

In summary, the contributions of this work are the following:
\begin{itemize}
\item we propose a novel methodology for online motion planning in partially unknown cluttered dynamic environments, \emph{requiring only the knowledge of the current position of obstacles}, rather than their velocities or full trajectories;
\item we discuss the assumptions to \emph{guarantee safe collision avoidance} with VO action pruning in MCTS;
\item we scale up the efficiency of MCTS to large action spaces (up to 60 actions) required for smooth trajectory generation in real robotic applications, introducing a VO-based methodology to prune unsafe actions and reduce required simulations;
\item we empirically demonstrate the efficacy of our methodology in environments with dense randomly moving obstacles, compared to state-of-the-art online planners (NMPC and DWA), and perform an ablation study to highlight benefits of VO in MCTS.
\end{itemize}


\section{Related Works}\label{sec:sota}
The problem of motion planning for mobile robots has been extensively studied in scientific literature \cite{dong2023review, antonyshyn2023multiple}. Nonetheless, this is still an open research area, given the demonstrated intractability of the problem in generic dynamic environments \cite{latombe2012robot}.
We classify motion planning algorithms into three main categories: reactive planners, look-ahead planners, and learning-based planners.

\emph{Reactive planners} compute only the next safe (i.e., collision-free) robot command, given the current configuration of the environmental map. Since they consider only the instantaneous situation, reactive planners are computationally efficient, hence the trajectory can be adapted at run time in the presence of dynamic obstacles and multiple moving agents. Main examples include Artificial Potential Fields (APF) \cite{ginesi2019dynamic, yao2020path, ginesi2021dynamic} and Velocity Obstacles (VO) \cite{fiorini1998motion,morasso2024planning}. A known issue with reactive planners, such as the APF method, is that they can get stuck in local minima \cite{fan2020improved} in case of specific configurations of obstacles and goals. This typically requires ad-hoc modifications of the standard reactive planning approach \cite{pan2021improved}. In the VO paradigm, a slight perturbation of the reactive planner may help escape local minima\footnote{Example implementation available at \url{https://gamma.cs.unc.edu/RVO2/}}; however, in cluttered environments this may badly affect the performance of the agent.

In complex maps, \emph{look-ahead planners} are more suitable to find a feasible path towards the goal, since they optimize the trajectory over a time horizon. Most popular examples include planners based on graph search to optimize the trajectory over a discretization of the environmental map, including Rapidly-exploring Random Trees (RRT) \cite{lukyanenko2023probabilistic} and A$^*$ \cite{erke2020improved}. Other look-ahead solutions include time-optimal planning \cite{foehn2021time}, Dynamic Window Approach (DWA) \cite{fox1997dynamic}, Non-linear Model Predictive Control (NMPC) \cite{liu2017path, piccinelli2023mpc} and MCTS \cite{gupta2022intention,Castellini_EAAI2021}.
Graph-based planners typically fail to scale in cluttered environments.  In such settings, the environment must be represented with very fine-grained discretization. This results in a prohibitively large search space. Furthermore, these approaches often require substantial computational resources in dynamic scenarios, where frequent re-planning is necessary. 
For this reason, look-ahead planners typically require prior knowledge of the trajectories of moving obstacles \cite{tordesillas2021mader}, which however requires unrealistic perfect communication or perception capabilities. In the context of crowd navigation, recent works \cite{gupta2022intention} integrates a probabilistic model of human behaviour into a partially observable MDP for online safe navigation. However, planning over a partially observable state space introduces additional computational complexity; moreover, different obstacles may require different trajectory models, which can only be approximated via learning algorithms \cite{li2020socially}, without any guarantee of safe collision avoidance when deployed in the real world. 

Recently, \emph{learning-based} planners exploiting DRL have been proposed to solve the motion planning problem in complex dynamic environments \cite{aradi2020survey, zhang2022receding}. 
In \cite{riviere2021neural}, authors combined MCTS with a neural network trained from expert demonstrations for non-cooperative robot navigation.
However, these methods typically require large training datasets and significant computational power for the training phase. Moreover, they do not provide guarantees in inference on generic maps, thus requiring specific approaches, e.g., posterior formal verification, for safe deployment on real robots \cite{amir2023verifying}.
In fact, combining the goal reward with the cost related to obstacle avoidance requires accurate multi-objective reward shaping, which may be sub-optimal and provides no guarantee when applied to real robotic systems \cite{hayes2022practical}.

To overcome the limitations of existing planners, we combine the benefits of look-ahead planning with MCTS and reactive planning with VO. 
It is important to note that our approach operates within a context where the robot lacks knowledge of obstacle trajectories and behaviors, precluding direct comparisons with some state-of-the-art techniques, such as \cite{tordesillas2021mader} (where obstacle trajectories are assumed to be known) and \cite{gupta2022intention} (where a pedestrian model is assumed to be known).
Instead, our assumptions emulate the realistic scenario where the robot moves in partially unknown environments in the presence of heterogenous obstacles (e.g., other robots or humans) with an upper bound on their maximum velocity, and can only rely on its sensors (e.g., LIDARs or cameras) to estimate the positions of other entities.
In addition, MCTS performs online planning, hence our method does not require learning and it cannot be compared to DRL approaches which require offline training.
We use VO to prune the robot's unsafe actions (i.e., command velocities).
In this way, our methodology can scale up to larger action spaces than the state-of-the-art \cite{shaobo2020collision} (up to 60 velocity actions in our case), required for application in realistic robotic settings.
Moreover, our algorithm is successful and efficient even in the presence of cluttered dynamic environments, where approaches based on map discretization typically fail \cite{andreychuk2022multi}.
Finally, differentl from existing methods for motion planning combining MCTS with efficient heuristics \cite{riviere2021neural}, our methodology does not require offline training or the availability of expert demonstrations.

\section{Background}\label{sec:background}

We now introduce the fundamentals of the VO paradigm and Monte Carlo planning for MDPs, which are the base of the methodology described in this paper. 

\subsection{Velocity Obstacles (VO)}

\begin{figure}
    \centering
    \includegraphics[scale=0.3]{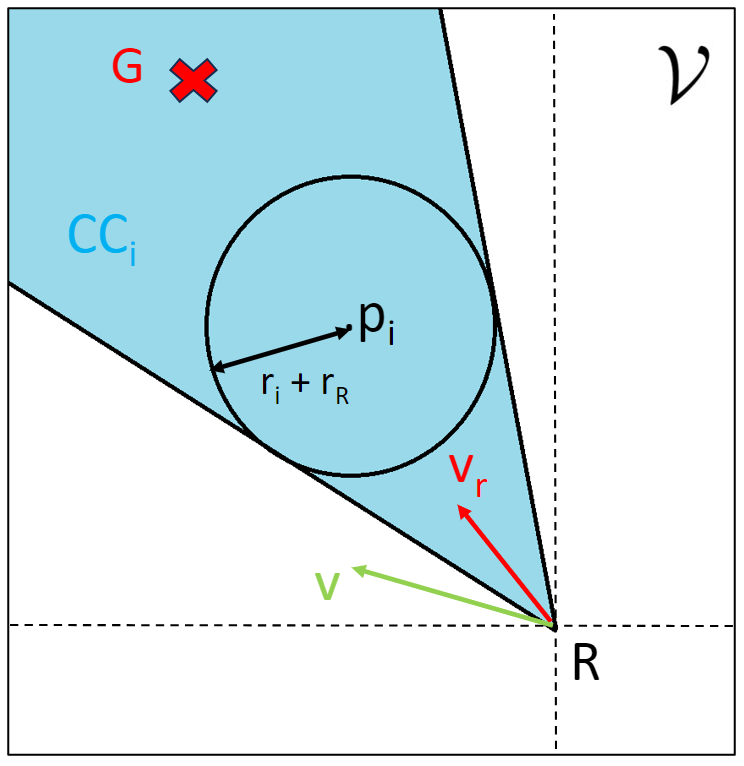}
    \caption{The velocity space for a robot moving in an environment with one obstacle. The blue region ($CC_i$) denotes the collision cone corresponding to the obstacle, a circle representing $\pazocal{B}(\bm{p}_i, r_i+r_R)$. Thus, $\bm{v}_r$ (the relative velocity of the robot to the obstacle) is infeasible, and another velocity (e.g., $v$) must be selected, such that $v \in \pazocal{V} \setminus CC_i$.}
    \label{fig:vo}
\end{figure}

In the classical VO setting, a robot $R$ must reach a target $G$ in an environment with $N$ obstacles.
Without loss of generality, we assume that the robot and the obstacles are spherical\footnote{More complex shapes can be considered, e.g., square obstacles \cite{van2008reciprocal}.}, with radii $r_R$ and $r_i, i
= 1, \ldots, N$, respectively.
At a given time step $\overline{t}$, the robot is at (vector) location $\bm{p}_R$, while the obstacles have positions $\bm{p}_i$ and velocity (vector) $\bm{v}_i$ (in our setting, the velocity of the obstacles is unknown).
Given the set $\pazocal{V}$ of admissible velocities for robot $R$, the VO paradigm is used to compute the set of collision-free velocities $\pazocal{V}_c \subseteq \pazocal{V}$. Specifically, for each $i$-th obstacle, we define a \emph{collision cone} $CC_i$ as:
\begin{equation}
\label{eq:cc}
    CC_i = \{\ \bm{v} \in \pazocal{V}\ |\ 
    \exists t>\overline{t} \text{ s.t. } \bm{p}_R(\overline{t}) + \bm{v}_r(t-\overline{t}) \cap \pazocal{B}(\bm{p}_i, r_R + r_i) \neq \emptyset \}
\end{equation}
\noindent
where $\pazocal{B}(\bm{p}_i, r_R + r_i)$ is the ball centered at $\bm{p}_i$ with radius $r_R + r_i$, and $\bm{v}_r = \bm{v} - \bm{v}_i$ is the relative velocity of the robot with respect to the obstacle.
We then define $\pazocal{V}_c = \pazocal{V} \setminus \bigcup_{i=1}^N CC_i$, the latter being the union of cones for every obstacle.
In Figure \ref{fig:vo}, we show the velocity space of the robot for a simple scenario with the collision cone for one obstacle.

\subsection{Markov Decision Processes}
A Markov Decision Process (MDP) \cite{bellman1957,puterman2014} is a tuple $M=\langle S,A,T,R,\gamma \rangle$, where
$S$ is a finite set of \textit{states} (e.g., robot and obstacle positions in the VO setting), $A$ is a finite set of \textit{actions} - we represent each action with its index, i.e., $A=\{ 1, \ldots, |A|\}$ (e.g., linear velocity and movement direction in the VO setting), $T: S\times A \rightarrow \mathcal{P}(S)$ is a stochastic or deterministic \textit{transition function} (e.g., the deterministic dynamics of the robot in the VO setting),  where $\mathcal{P}(E)$ denotes the space of probability distributions over the finite set $E$, therefore $T(s,a,s')$ indicates the probability of reaching the state $s'\in S$ after executing $a\in A$ in  $s \in S$,
$R: S \times A \times S \rightarrow [-R_{max},R_{max}]$ is a bounded stochastic \textit{reward function} (e.g., a function that rewards the robot if it gets close to or reaches the goal avoiding the obstacles, in the VO setting), and
$\gamma \in [0,1)$ is a \textit{discount factor}. The set of stochastic policies for $M$ is $\Pi = \{ \pi: S \rightarrow \mathcal{P}(A) \}$. In the VO setting used in this paper, a policy is a function that suggests the linear velocity and direction of the movement given the current position of the robot and the obstacles. Given an MDP $M$ and a policy $\pi$ we can compute state values $V_M^\pi(s), s \in S$, namely, the expected value acquired by $\pi$ from $s$; and action values $Q_M(s,a), s \in S, a \in A$, namely, the expected value acquired by $\pi$ when action $a$ is performed from state $s$. To evaluate the performance of a policy $\pi$ in an MDP $M$, called $\rho(\pi,M)$ in the following, we compute its expected return (i.e., its value) in the initial state $s_0$, namely, $\rho(\pi,M)=V_M^\pi(s_0)$. The goal of MDP solvers \citep{Russell2020,Sutton2018} is to compute optimal policies, namely, policies having maximal values (i.e., expected return) in all their states. 

\subsection{Monte Carlo Tree Search}
MCTS \cite{chaslot2008,Browne2012} is an online solver, namely, it computes the optimal policy only for the current state of the agent. 
In particular, given the current state of the agent, MCTS first generates, in a sample-efficient way, a Monte Carlo tree rooted in the current state of the agent. In this way, it estimates the Q-values (i.e., action values) for that state. Then, it uses these estimates to select the best action. A certain number $m\in \N$ of simulations are performed using, at each step, Upper Confidence Bound applied to Trees (UCT) \cite{Auer2002,Kocsis2006} (inside the tree) or a rollout policy (from a leaf to the end of the simulation) to select the action. The transition model (or an equivalent simulator) is used to perform the step from one state to the next. Simulations allow updating two node statistics, namely, the average discounted return~$Q(s,a)$ obtained by selecting action $a$ from state $s$, and the number of times~$N(s,a)$ action $a$ was selected from state~$s$. UCT extends UCB1 \cite{Auer2002} to sequential decisions and allows to balance exploration and exploitation in the simulation steps performed inside the tree, and to find the optimal action as $m$ tends to infinity. Given the average return $\bar Q_{a, T_a(t)}$ of each action $a\in A$ after $t$ simulations, where $T_a(t)$ is the number of times action $a$ has been selected up to simulation $t$, UCT selects the action with the best upper confidence bound. In other words, the index of the action selected at the $t$-th visit of a node is $I_t = \argmax_{a \in 1, \ldots, |A|}\ \bar Q_{a, T_a(t)} + 2C_p \sqrt{\frac{\ln (t-1)}{T_a(t-1)}}$, with appropriate constant $C_p>0$. When all $m$ simulations are performed, the action $a$ with maximum average return (i.e., Q-value)~$\bar Q_{a, T_a(m)}$ in the root is executed in the real environment.

\section{Methodology}\label{sec:met}
We consider a 2D motion planning scenario, where the robot has only access to the current position of obstacles, their maximum speed and radius, hence requiring no strict communication capabilities. It is crucial to note that no specific knowledge is available about obstacle behaviors. This replicates a realistic robotic setting where an agent can rely only on its sensors (e.g., LIDARs or cameras) to estimate the current state of the surrounding environment.
We define the problem with $N$ dynamic obstacles and a goal in position $\bm{p}_G$ as a MDP with state $S = \{\langle \bm{p}_R, \bm{v}_R, \pazocal{P}_o, \pazocal{R}_o, \pazocal{V}_{max,o}\rangle\}$ and $A = \pazocal{V}$ (i.e., admissible velocities for the robot), where $\pazocal{P}_o = \left\{\bm{p}_i\right\}_{i=1}^N$ and $\pazocal{R}_o = \left\{r_i\right\}_{i=1}^N$ denote the list of obstacles positions and radii, respectively; $\pazocal{V}_{max,o}$ is the list of maximum speeds of the obstacles.  
Notice that the state space is in general continuous, because the robot can reach all possible positions in the environment, and the action space is also continuous (we will discretize it later to use it in MCTS).

The reward for a given tuple $\langle s,a,s'\rangle$ (state, action and next state, respectively) is modeled as:
\begin{equation}\label{eq:rew}
    R(s,a,s') = 
    \begin{cases}
    R_h &\text{if at } s',\ ||\bm{p}_G - \bm{p}_R||_2 < r_R\\
    -R_h &\text{if at } s',\ \pazocal{B}(\bm{p}_R, r_R) \text{ out of } W_{lim}\\
    -R_h &\text{if at } s',\ \exists i \ \text{s.t.} \ ||\bm{p}_i - \bm{p}_R||_2 < r_R+r_i\\
    -\frac{||\bm{p}_R - \bm{p}_G||_2}{d_{max}} &\text{otherwise}
    \end{cases}
\end{equation}
where $R_h$ is a high reward value\footnote{$R_h=100$ in our experiments.}, $\pazocal{B}(\bm{p}_R, r_R)$ is the ball centered at $\bm{p}_R$ with radius $r_R$, $W_{lim}$ denotes the limits of the workspace and $d_{max}$ is the maximum distance to the goal in the given map (e.g., in a square 2D map, it is the diagonal dimension).
Equation \eqref{eq:rew}, rewards goal reaching (first line), penalizes going out of workspace bounds (second line) or hitting obstacles (third line), and penalizes the normalized distance to the goal (last line). 
The transition map $T$ is deterministic.

The main idea of the proposed methodology is to introduce the VO constraint in the simulation process performed by MCTS to estimate action values. This can improve the efficiency of the simulation process, allowing only exploration of $\pazocal{V}_c \subseteq \pazocal{V}$ (i.e., safe collision-free velocities). 


\subsection{Action space discretization}\label{sec:discretization}
We express each velocity $\bm{v} \in \pazocal{V}$ as a tuple $\bm{v} = \langle v, \alpha \rangle$, where $v$ is the module of the velocity and $\alpha$ is the heading angle in radians.
We assume that the physical constraints of the robot impose a maximum velocity module $v_{max}$ and a maximum angular velocity $\omega_{max}$. Thus, at each time step $t_s$, where the robot has heading angle $\alpha_0$, the action space can be expressed as $A = \{\langle v, \alpha \rangle \ |\ 0 \leq v \leq v_{max}, \alpha_0 - \omega_{max}t_s \leq \alpha \leq \alpha_0 + \omega_{max}t_s \}$.
We then obtain the action space for MCTS by discretizing $v$ and $\alpha$ within their respective ranges\footnote{We discretize $v$ into 5 equally spaced values and $\alpha$ into 11 equally spaced values} and considering all possible combinations of them. We also add actions in the form $\langle 0, \alpha \rangle$, to allow in-place rotation.

\subsection{Integration of VO into MCTS}
\begin{algorithm}
    \caption{Compute $\pazocal{V}_c$}\label{alg:cc}
    \begin{algorithmic}[1]
    \Function{Compute\_Vel}{$r_R, \bm{p}_R, \pazocal{R}_o, \pazocal{P}_o, v_{max}, \alpha_0, \pazocal{V}_{max,o}, t_s$}
      \State\textit{\% Set of feasible angles within kinematic limits}
      \State $\pazocal{A}_c \leftarrow \{ \alpha \ |\ \alpha_0 - \omega_{max}t_s \leq \alpha \leq \alpha_0 + \omega_{max}t_s\}$ 
      \For{$\bm{p}_i \in \pazocal{P}_o \land r_i \in \pazocal{R}_o \land v_{max,i} \in \pazocal{V}_{max,o}$}
      \State $r_1 \leftarrow v_{max}t_s$ 
      \State $r_2 \leftarrow r_i+r_R+v_{max,i}t_s$ 
      \If{$\pazocal{B}(\bm{p}_R, r_1) \cap \pazocal{B}(\bm{p}_i, r_2) \neq \emptyset \land \bm{p}_R \notin \pazocal{B}(\bm{p}_i, r_2)$}
      \State $\langle \alpha_1, \alpha_2 \rangle \leftarrow$ tangent angles from $\bm{p}_R$ to $\pazocal{B}(\bm{p}_i, r_2)$
      \State $\pazocal{A}_c \leftarrow \pazocal{A}_c \setminus [\alpha_1, \alpha_2]$
      \ElsIf{$\bm{p}_R \in \pazocal{B}(\bm{p}_i, r_2)$}
      \State $\pazocal{A}_c \leftarrow \emptyset$
      \State \textbf{break}
      \EndIf
      \EndFor
      \If{$\pazocal{A}_c\neq\emptyset$} 
        \State$V_c \leftarrow [0, v_{max}]$
      \Else
        \State$V_c \leftarrow \{0\}$
      \EndIf
    \EndFunction
    \Return $\pazocal{V}_c = V_c \times \pazocal{A}_c$
    \end{algorithmic}
\end{algorithm}
We introduce VO in two phases of MCTS, namely, Monte Carlo tree exploration, where UCT selects actions in simulation steps performed inside the Monte Carlo tree (in the following, this method will be called MCTS\_VO\_TREE) and rollout, where a random policy selects actions in simulation steps performed out of the Monte Carlo tree (this method is called MCTS\_VO\_ROLLOUT in the following). 
In both cases, we build collision cones as in Eq. \eqref{eq:cc}, but considering only collisions happening within the duration of a time step $t_s$ in the simulation \cite{claes2012collision}, which is the discretization time step to compute the transition between subsequent states in the MDP.
To reduce the computational complexity in finding the set of collision-free velocities $\pazocal{V}_c$, we consider the worst-case scenario where the module of the robot's velocity is $v_{max}$. In fact, if the robot cannot reach the obstacle at $v_{max}$, even lower velocities will be safe. 
Similarly, we assume that obstacles move at velocity with module $v_{max,i} \in \pazocal{V}_{max,o}$. 
In this way, the agent only needs position information about the obstacles (which are easier to estimate from sensors) instead of their velocity \cite{xu2020collision, du2020improving}. In this conservative scenario, computing the set of safe (i.e., collision-free) velocities is equivalent to compute the set of safe heading angles $\pazocal{A}_c$ as indicated in Lines 8-9 of Algorithm \ref{alg:cc}. 
When assessing potential collisions, Algorithm \ref{alg:cc} examines intersections between the collision cones formed by the agent and obstacles, represented by the intersection of balls $(\pazocal{B}(\bm{p}_R, r_1) \cap \pazocal{B}(\bm{p}_i, r_2))$ (Lines 5-7).
When the intersection is non-empty, we compute lines from $\bm{p}_R$ to the intersections between the collision cones \footnote{In case of walls, we model them as segments and compute lines from $\bm{p}_R$ to their extremes.} (Line 8 and Figure \ref{fig:vo_int}).
Angles within these lines denote directions of potential collision at maximum speed. Consequently, outside angles are deemed safe for the agent to navigate at maximum speed (Line 14). In cases where no safe angles are available within the environment, the only viable option for the robot is to remain stationary (Line 16).

\subsubsection{VO in UCT}
\begin{algorithm}
    \caption{UCT expansion with VO constraints}\label{alg:expand}
    \begin{algorithmic}[1]  
    \Require radius of robot $r_R$, list of radii of obstacles $\pazocal{R}_o$, position of robot $\bm{p}_R$, list of positions of obstacles $\pazocal{P}_o$, current heading angle $\alpha_0$, maximum agent linear and angular velocity modules $v_{max}, \omega_{max}$, maximum obstacle velocities $\pazocal{V}_{max,o}$, simulation time step $t_s$
    \Function{Expand}{$n = \langle s, a \rangle$}
        \State $\pazocal{V}_c = V_c \times \pazocal{A}_c \leftarrow$\texttt{COMPUTE\_VEL}($r_R, \bm{p}_R, \pazocal{R}_o, \pazocal{P}_o, v_{max}, \alpha_0$)
        \State Choose new $a' \in \pazocal{V}_c$ with UCT \cite{surveymcts}
        \State $s' \sim T(s,a')$
        \State Add new child $n' = \langle s, a' \rangle$ to $n$
        \State\Return $n'$
    \EndFunction
    \end{algorithmic}
\end{algorithm}
\begin{algorithm}
    \caption{Rollout}\label{alg:vor}
    \begin{algorithmic}[1]
    \Require radius of robot $r_R$, list of radii of obstacles $\pazocal{R}_o$, position of robot $\bm{p}_R$, list of positions of obstacles $\pazocal{P}_o$, current heading angle $\alpha_0$, maximum agent linear and angular velocity modules $v_{max}, \omega_{max}$, maximum obstacle velocities $\pazocal{V}_{max,o}$, simulation time step $t_s$, uniform selection threshold $\epsilon_0$, heading direction to the goal $\alpha_G$, 
    \Function{Rollout}{state $s$}
        \While{$s$ is not terminal}
            \If{Using VO}
            \State $\pazocal{V}_c = V_c \times \pazocal{A}_c \leftarrow $\texttt{COMPUTE\_VEL}($r_R, \bm{p}_R, \pazocal{R}_o, \pazocal{P}_o, v_{max}, \alpha_0$)
            \Else
            \State $\pazocal{V}_c = V_c \times \pazocal{A}_c \leftarrow$ space of feasible actions
            \EndIf
            \State choose $\epsilon\in[0,1]$ uniformly random
            \If{$\epsilon\le\epsilon_0$}
            \State choose $a \in \pazocal{V}_c$ uniformly  random
            \Else
            \State choose $\alpha \in [\alpha_G - \delta, \alpha_G + \delta] \cap \pazocal{A}_c$ uniformly random
            \State choose $v \in V_c$ uniformly random
            \State $a \leftarrow [v, \alpha]$
            \EndIf
            \State $s' \sim T(s,a)$
        \EndWhile
    \EndFunction
    \end{algorithmic}
\end{algorithm}
\label{sec:met_uct}
It concerns simulation steps taken inside the Monte Carlo tree, namely, the first steps performed near the robot's current position, to realize safe collision avoidance in the short term of the trajectory execution.
Algorithm \ref{alg:expand} shows the pseudo-code implementation of branch expansion in UCT based on VO\footnote{For the full UCT algorithm, please refer to \cite{surveymcts}.}.
Considering the maximum possible relative velocity between the agent and other robots/obstacles, at each step of UCT we compute the set $\pazocal{V}_c$ of velocity vectors that do not lead to collisions (Line 2 in Algorithm \ref{alg:expand}, using Algorithm \ref{alg:cc}), i.e., the set of vectors such that the resulting relative velocity $\bm{v}_r$ does not belong to any of the collision cones in Equation \eqref{eq:cc} within $t_s$. In this way, we prune away all unsafe actions, reducing the size of the action space and then making simulations more efficient. 

We remark that the UCT phase is where the agent selects the optimal action to execute in the real environment.
Hence, VO in this phase is crucial for safe mobile robot navigation.
In an ablation study (see next section), we found that introducing VO in the UCT phase (algorithm MCTS\_VO\_TREE in Section \ref{sec:exp}) yields the most promising results compared to other implementations. 
Therefore, we designate this configuration as our benchmark for comparison with the baselines.

\subsubsection{VO in Rollout}\label{sec:vo_rollout}
The rollout phase of MCTS is known to be computationally demanding, hence it is important to design suitable heuristics to guide action selection \cite{meli2024learning, riviere2021neural} and shield undesired or unsafe actions \cite{Mazzi2023b}. 
As a rollout policy (Algorithm \ref{alg:vor}), we use an heuristic to encourage the robot to move in the direction of the goal but also ensure convergence guarantees. Specifically, if the direction between the robot and the goal corresponds to angle $\alpha_G$, we sample angles within $\pazocal{A}_c$ with uniform distribution in $[\alpha_G - \delta, \alpha_G + \delta]$, as in Line 9, being $\delta$ an empirical parameter ($\delta = $ \SI{1}{rad} in this paper).
However, since the robot may get stuck due to obstacles on the way to the goal, we implement an $\epsilon$-greedy strategy (Line 6) to follow this heuristic with probability $1-\epsilon_0$.
The rollout policy with the VO constraints (Line 4 of Algorithm \ref{alg:vor}), instead of sampling from all the space, samples only safe angles within $\pazocal{A}_c$ by building collision cones and performing action pruning similarly to Algorithm \ref{alg:expand}.

\subsection{Assumptions for safe collision avoidance}\label{sec:ass}
\begin{figure}
    \centering
    \subfloat[]{\includegraphics[width=0.16\textwidth]{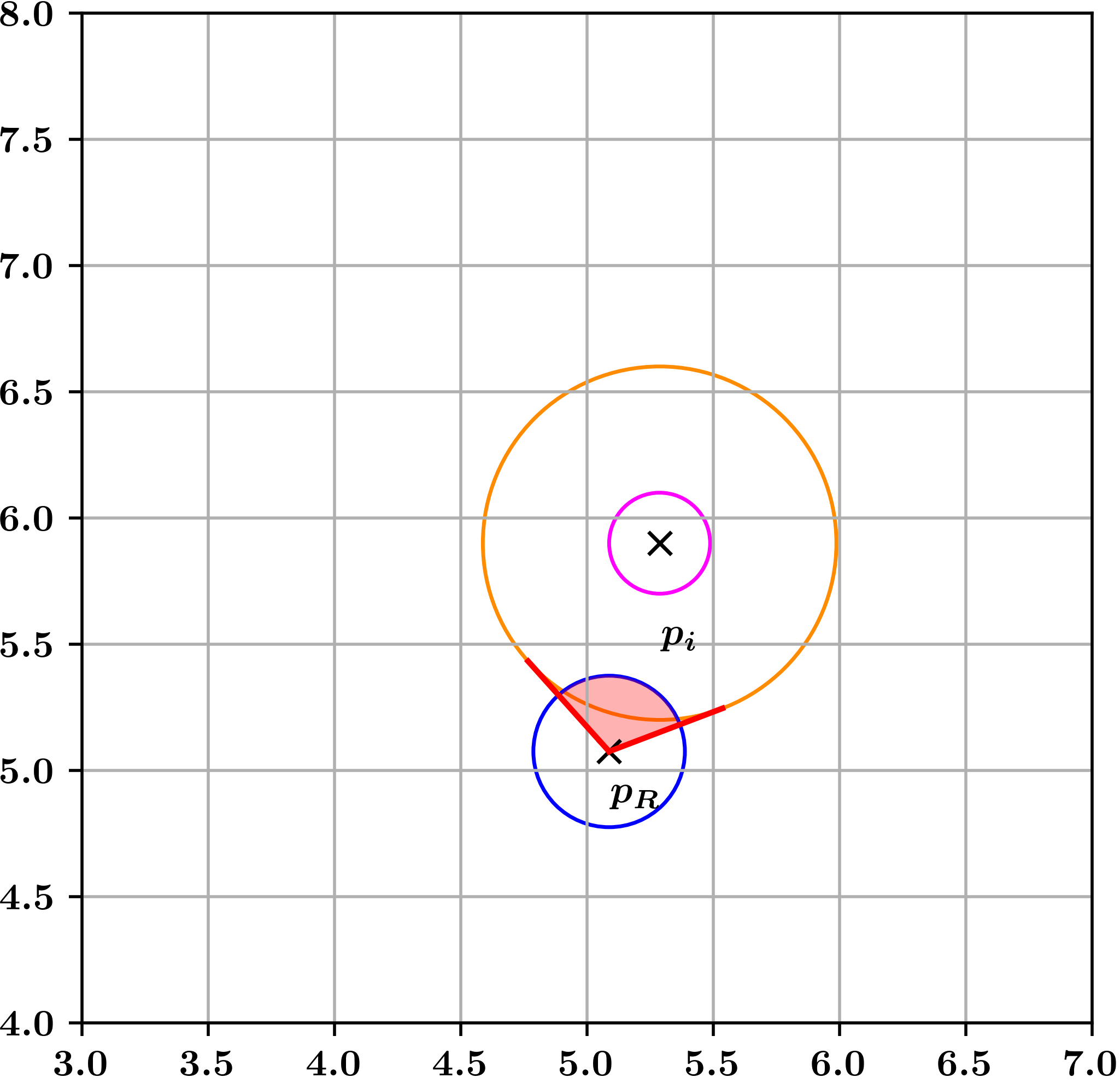}\label{fig:vo_int}}
    \hspace{0.05\textwidth}
    \subfloat[]{\includegraphics[width=0.16\textwidth]{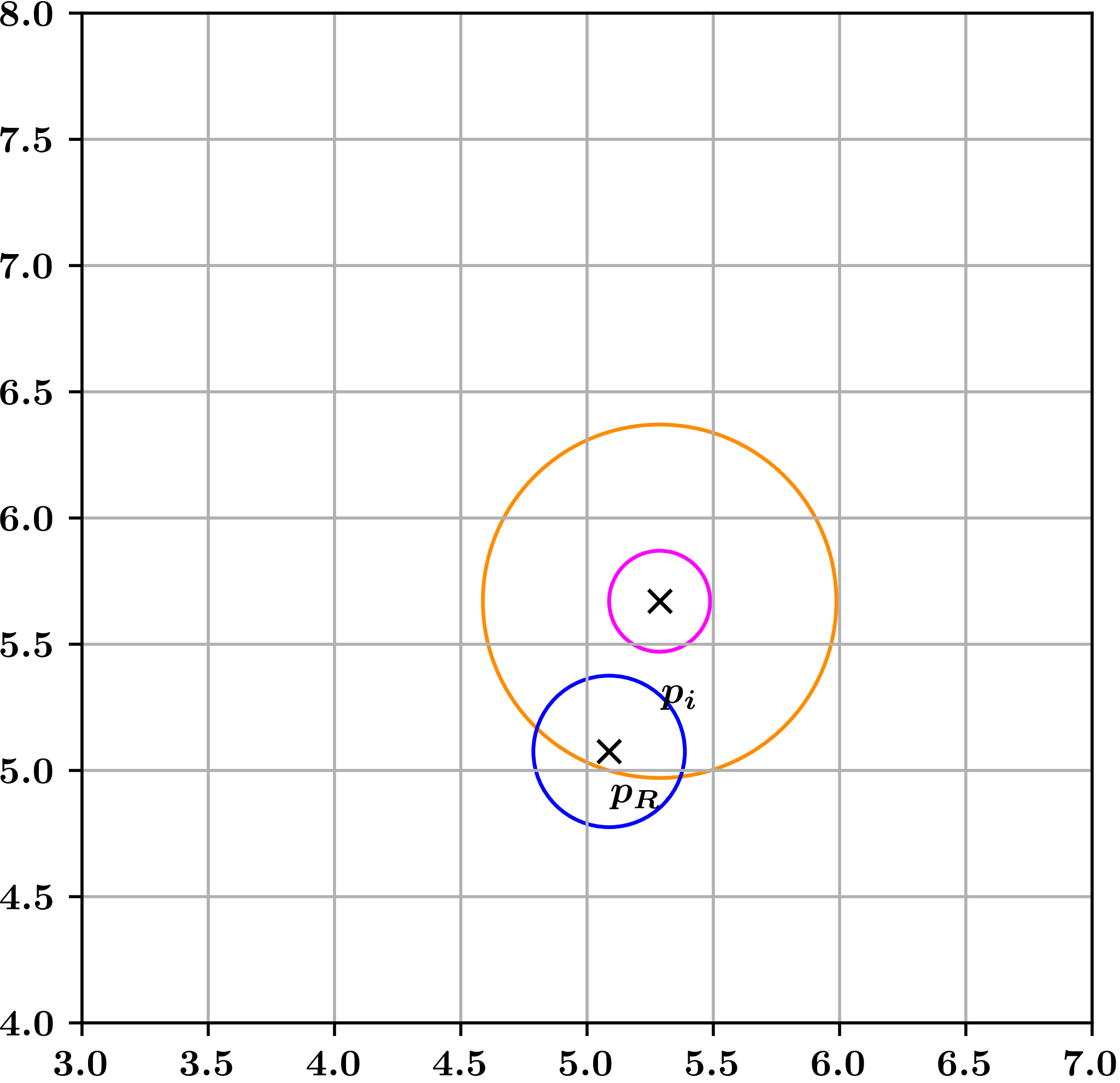}\label{fig:vo_comp}}
    \caption{a) The robot ($\bm{p}_R$) is out of the extended ball $\pazocal{B}(\bm{p}_i, r_2)$ (yellow circle); b) The robot is inside the extended ball, but still not colliding with the physical ball of the obstacle $\pazocal{B}(\bm{p}_i, r_i)$ (pink circle). Blue circle: $\pazocal{B}(\bm{p}_R, r_1)$; red cone: $CC_i$ delimited by tangent angles $[\alpha_1, \alpha_2]$ (Algorithm \ref{alg:cc}, Line 8).}
\end{figure}
Introducing VO in MCTS allows to prune colliding actions (velocities), hence improving the planning efficiency and reducing the rate of collision with obstacles.
In this paper, we assume minimal knowledge about the environment (i.e., only positions and maximum possible velocities of obstacles) for the best adherence to real-world partially observable robotic settings, showing the significant advantages of our methodology experimentally in the next section.
However, the algorithm cannot \emph{always guarantee} safe collision avoidance because collisions also depend on the behavior of dynamic obstacles.
We now discuss in more detail our assumptions and necessary modifications to them in order to achieve formal guarantees, showing their feasibility in realistic robotic settings.

\subsubsection{Knowledge of $\pazocal{V}_{max,o}, \pazocal{P}_o$}
Our methodology assumes that the positions and maximum velocities of obstacles are available to the agent at each step of MCTS computation. In general, obstacle positions can be estimated with standard robotic sensors, e.g., LIDARs, while estimating actual velocities is more challenging due to additional noise in the computation \cite{thakur2024l3d}. On the contrary, estimating the \emph{maximum} possible velocities is easier, since they can be derived from available rough prior information (e.g., crowd models \cite{karamouzas2014universal, luo2018porca} or other robots' specifications). 
Moreover, when computing safe velocities in Algorithm \ref{alg:cc}, we can enlarge $r_{1,2}$ (Lines 5-6) with safety bounds, in order to incorporate the level of confidence of sensors and uncertainties about $\pazocal{V}_{max,o}$. Note that, in case of large uncertainty, $\pazocal{A}_c = \emptyset$ in Algorithm \ref{alg:cc}. In this case, our algorithm is still able to take a collision-free action, which is remaining stationary (Line 16).

\subsubsection{Unknown obstacle trajectories} 
Starting from a configuration where the center of the robot is outside the extended circumference $\bm{p}_R \not\in \pazocal{B}(\bm{p}_i, r_2)$, see Figure \ref{fig:vo_int}, we can guarantee collision avoidance one step ahead, using Algorithm \ref{alg:cc}.
However, there may be situations where Algorithm \ref{alg:cc} cannot compute tangent angles to the obstacle (Line 8), i.e., when $\bm{p}_R \in \pazocal{B}(\bm{p}_i, r_2)$\footnote{This situation does not necessarily entail collision. Indeed, the actual collision occurs when the robot intersects with the physical radius of the obstacle, i.e., $\pazocal{B}(\bm{p}_R, r_R) \cap \pazocal{B}(\bm{p}_i, r_i)$.} (see Figure \ref{fig:vo_comp}). This typically occurs when the obstacle moves too close towards the robot. Such situation cannot be taken into account by our algorithm, since we do not assume any prior knowledge about the intended trajectories of obstacles. 
Even in this condition, our algorithm guarantees that the robot does not take colliding actions, commanding null velocity (Lines 11 and 16 in Algorithm \ref{alg:cc}) but the obstacle could hit the robot. This situation should, however, not occur frequently in practical settings, e.g., in crowd navigation people tend to preserve a minimum distance to minimize the risk of collisions \cite{karamouzas2014universal}.
Hence, our approach works in standard settings. We will consider adversarial situations in which \emph{malicious} obstacles \emph{deliberately} try to collide with the robot in future works. 

\subsubsection{Simulation time}\label{sec:ts_condition}
In Algorithm \ref{alg:cc}, safe velocities are computed assuming that the obstacles move at $\pazocal{V}_{max,o}$ during the simulation time step $t_s$. Hence, collision avoidance is only guaranteed if the planning time step (i.e., the time required by MCTS to compute the optimal action to be executed) is lower than $t_s$ (i.e., the step time in MCTS, needed to compute the action space $A$ in Section \ref{sec:discretization}).
In our experiments, we will show that VO action pruning significantly increases the efficiency of MCTS, thus realizing this requirement for safe collision avoidance.

\section{Experimental results}\label{sec:exp}
We evaluate each algorithm with a number of MCTS simulations ranging in [10, 400]\footnote{Above 400 simulations per step, we did not notice significant variations in the results.} and investigate the performance as descibed in Section \ref{sec:perf_mes} in terms of \emph{discounted return}, \emph{collision rate}, and \emph{computational time per step}. To account for the stochasticity of MCTS, we run 50 tests for each number of simulations. In the action space, we consider 60 actions, which combine 5 different velocity modules up to robot's $v_{max}$ and 12 different heading angles among the feasible ones.
The discount factor for MDP is chosen empirically as $\gamma=0.7$.
In the rollout phase, we empirically set $\epsilon_0 = 0.2$ in Algorithm \ref{alg:vor}. The maximum number of allowed steps in simulation is set to 100 (i.e tree depth + rollout steps).

All experiments are run with Python 3.10 on a PC with Processor Intel(R) Core(TM) i5-13600KF CPU, 64GB Ram running Ubuntu 22.04.2 LTS. 
The code is available in the supplementary material.

\subsection{Domain}
\begin{figure}
    \centering
    \subfloat[]{\includegraphics[width=0.24\textwidth]{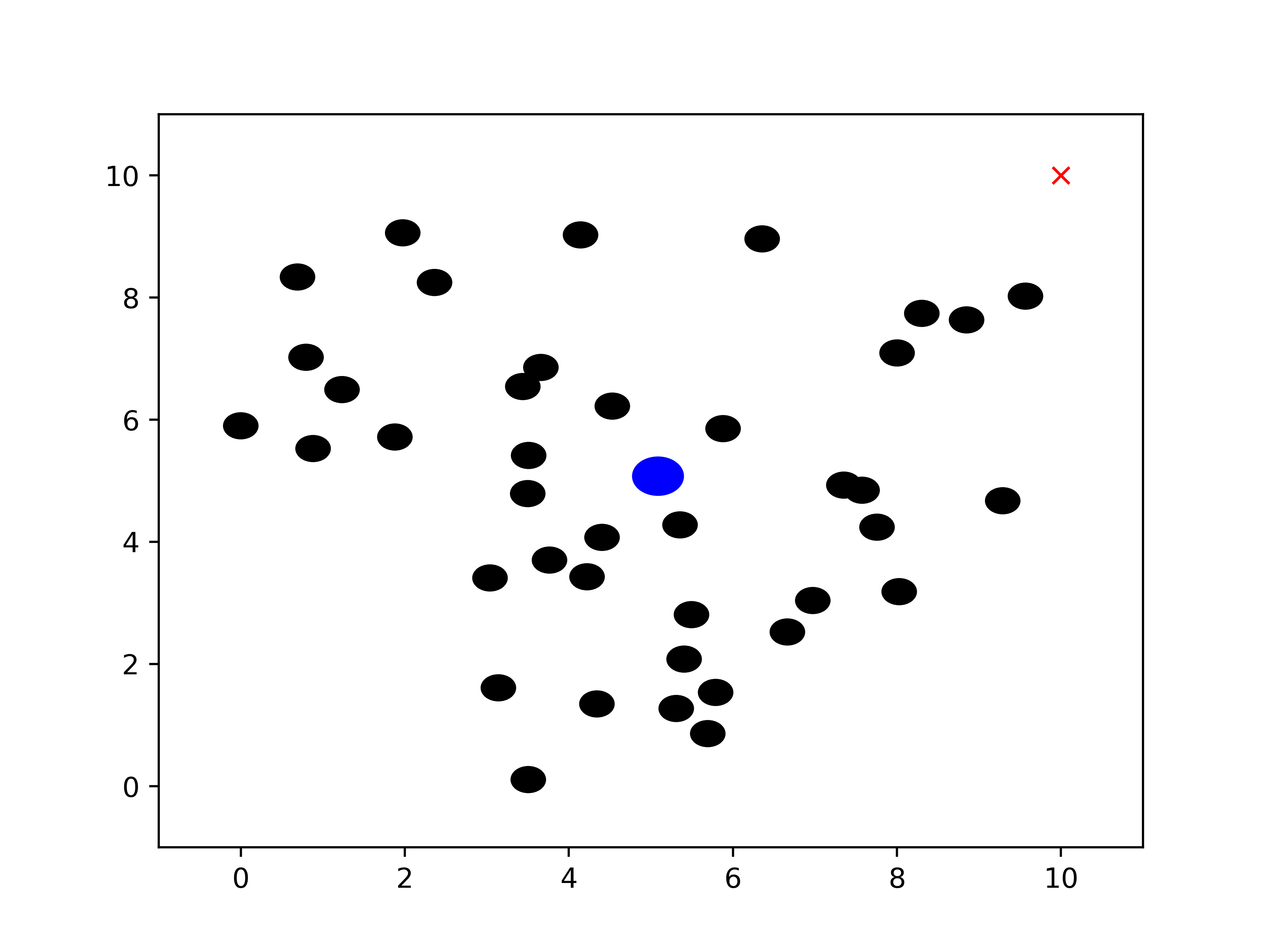}\label{fig:env}}
    \subfloat[]{\includegraphics[width=0.24\textwidth]{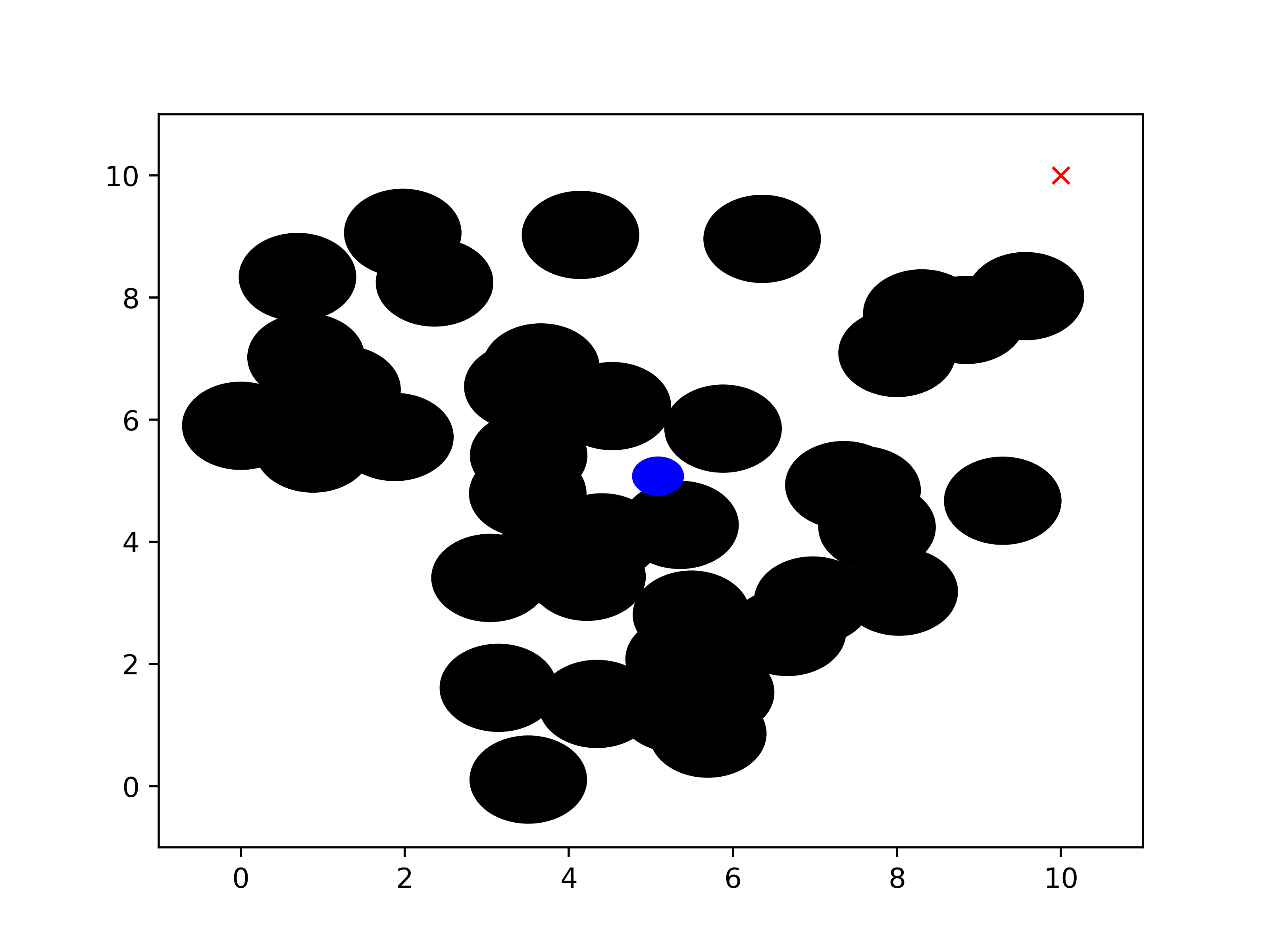}\label{fig:large_env}}
    \caption{Snapshot of the actual environment with obstacle radius $r_i = \SI{0.2}{m}$ and robot radius $r_R = \SI{0.3}{m}$ (a); and radii $r_1, r_2$ as of Algorithm \ref{alg:cc}. Blue circle: agent. Black circles: dynamic obstacles. Red cross: goal.}
\end{figure}
We evaluated our methodology in the simulated map depicted in Figure \ref{fig:env}, consisting of a $10\times 10$\si{m} workspace containing 40 dynamic obstacles. 
 
For dynamic obstacles modeling, we set all maximum velocities in $\pazocal{V}_{max,o}$ as $v_{max,o} = \SI{0.2}{m/s}$.
At each time step, dynamic obstacles move towards randomly predefined goals on the map, with additive noise\footnote{We also successfully tested our methodology against benchmark deterministic trajectories for obstacles, e.g., trefoils proposed in \cite{tordesillas2021mader}. Due to page limits, the results are reported in the supplementary materials.}.
Specifically, at each time step the velocity is sampled uniformly in $[-\frac{v_{max,o}}{2}, \frac{v_{max,o}}{2}]$; then the heading angle is the obstacle's goal direction, plus uniform noise in $[-0.05, 0.05]$.
For robot modeling, we set realistic values $v_{max} = \SI{0.3}{m/s}$ and $\omega_{max} = \SI{1.9}{rad/s}$ from the datasheet of Turtlebot 4 by Clearpath Robotics, which is an established open-source mobile robotic platform for research and education \footnote{\url{https://clearpathrobotics.com/turtlebot-4/}}.
We performed our experiments across 50 distinct scenarios, randomly varying the initial positions of the obstacles, to thoroughly assess the performance of our motion planning strategy. 
In all scenarios, the robot has radius $r_R = $ \SI{0.3}{m}, while each $i$-th obstacle has radius $r_i = $ \SI{0.2}{m}.
The chosen setup deliberately represents a cluttered environment, constituting a challenging planning context that requires the adoption of an efficient and safe look-ahead planning strategy. 
In particular, the considered setting prevents the use of a standard map discretization approach, as it would prohibitively increase the computational complexity. 
This is evidenced in Figure \ref{fig:large_env}, where the VO-enlarged radii of obstacles and robot are shown (Lines 5-6 in Algorithm \ref{alg:cc}).

The discrete time step of motion is chosen empirically as $t_s = $ \SI{1}{s}, but it can be adapted depending on the specific map configuration and task needs. 

\subsection{Algorithms and Baselines}\label{subsec:algo}
We assess the effectiveness of various look-ahead planners, operating on the premise of lacking knowledge about obstacle trajectory and without the need for prior offline training. The algorithms evaluated in our experiments are listed in the following:
\begin{itemize}
    \item \emph{MCTS\_VO\_TREE}: it is the algorithm introducing the VO constraint inside MCTS only in the simulation steps performed inside the tree (see Section \ref{sec:met_uct}). This is the version of our approach showing the best tradeoff between performance and computational time (see Section \ref{sec:ablation});
    \item \emph{MCTS\_VO\_ROLLOUT}: it is the algorithm introducing the VO constraint inside MCTS only in the simulation steps performed during the rollout phase (see Section \ref{sec:vo_rollout});
    \item \emph{MCTS\_VO\_2}: it is the algorithm introducing the VO constraint both inside the Monte Carlo tree and in the rollout phase;
    \item \emph{MCTS}: standard implementation of MCTS (without VO). This is used as a baseline to show the improvement achieved introducing VO within MCTS;
    \item \emph{Non-linear Model Predictive Control (NMPC)}: a state-of-the-art planning algorithm based on nonlinear model predictive control and an established methodology for optimal motion planning \cite{tordesillas2021mader}. In particular, it is a look-ahead planner which solves a constrained optimization problem over an horizon $\tau$, where the reward components in Equation \eqref{eq:rew} are converted into cost components (i.e the sign is inverted). 
    \item \emph{Dynamic Window Approach (DWA)}: a standard methodology for efficient optimal planning in dynamic environments \cite{matsuzaki2021dynamic}.
    \item \emph{VO\_PLANNER}: a reactive planner based on VO and informed random exploration of angles of velocities towards the goal, following Lines 3-10 of Algorithm \ref{alg:vor}. This baseline is based on VO, hence provides the same theoretical guarantees as our methodology. However, as shown in the following experiments, it does not perform look-ahead planning, thus resulting in poorer performance in our challenging map.
\end{itemize}

\subsection{Performance measures} \label{sec:perf_mes}
To evaluate the performance of the algorithms, we use the three measures defined in the following:
\begin{itemize}
    \item \emph{Discounted Return} ($\rho$): it is the discounted sum of all rewards obtained during the course of a trajectory, i.e., $\rho=\sum_{k = 0}^H \gamma^{k-1}r_{k}$, where $H$ is the total number of steps in the trajectory. It indicates the quality of the trajectory and the travel distance, since at each step the agent cumulates a small negative reward at each step. This measure must be maximized.
    \item \emph{Collision rate} ($\eta$): it is the ratio between the number of episodes in which the agent collides and the total number of episodes performed, i.e., 
    $\eta=\frac{\sum_{e = 1}^{n\_exp} \mathbb{1}_{collides}}{n\_exp}$. This measure must be minimized, as it quantifies the reliability of the planner across different environment settings.
    \item \emph{Planning time per step} ($t_{plan}$): it is the average computational time taken by the planner to compute a planning step. It is useful to evaluate the applicability of each planning algorithm to real-world robotic setting, where the time available for deciding the next action is limited. This measure must be minimized and kept below $t_s$ (see Section \ref{sec:ts_condition}).
\end{itemize}

\subsection{Comparison with baselines}\label{sec:comparison}
\begin{figure*}
    \centering
    \subfloat[Collision Rate ($\eta$)]{\includegraphics[width=0.33\textwidth]{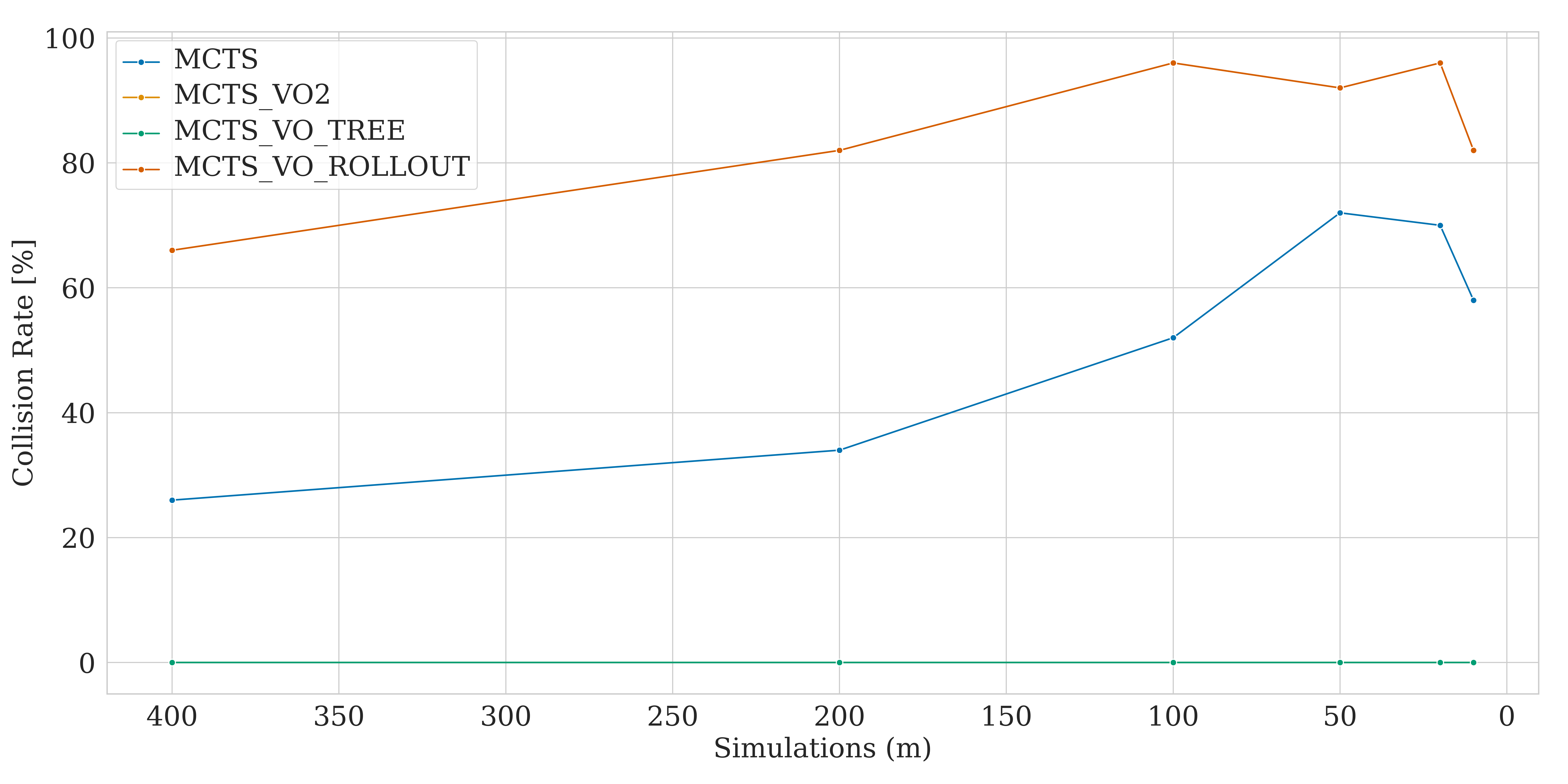}\label{fig:1confronto_coll}}
    \subfloat[Discounted Return ($\rho$, mean $\pm$ std. dev.)]{\includegraphics[width=0.33\textwidth]{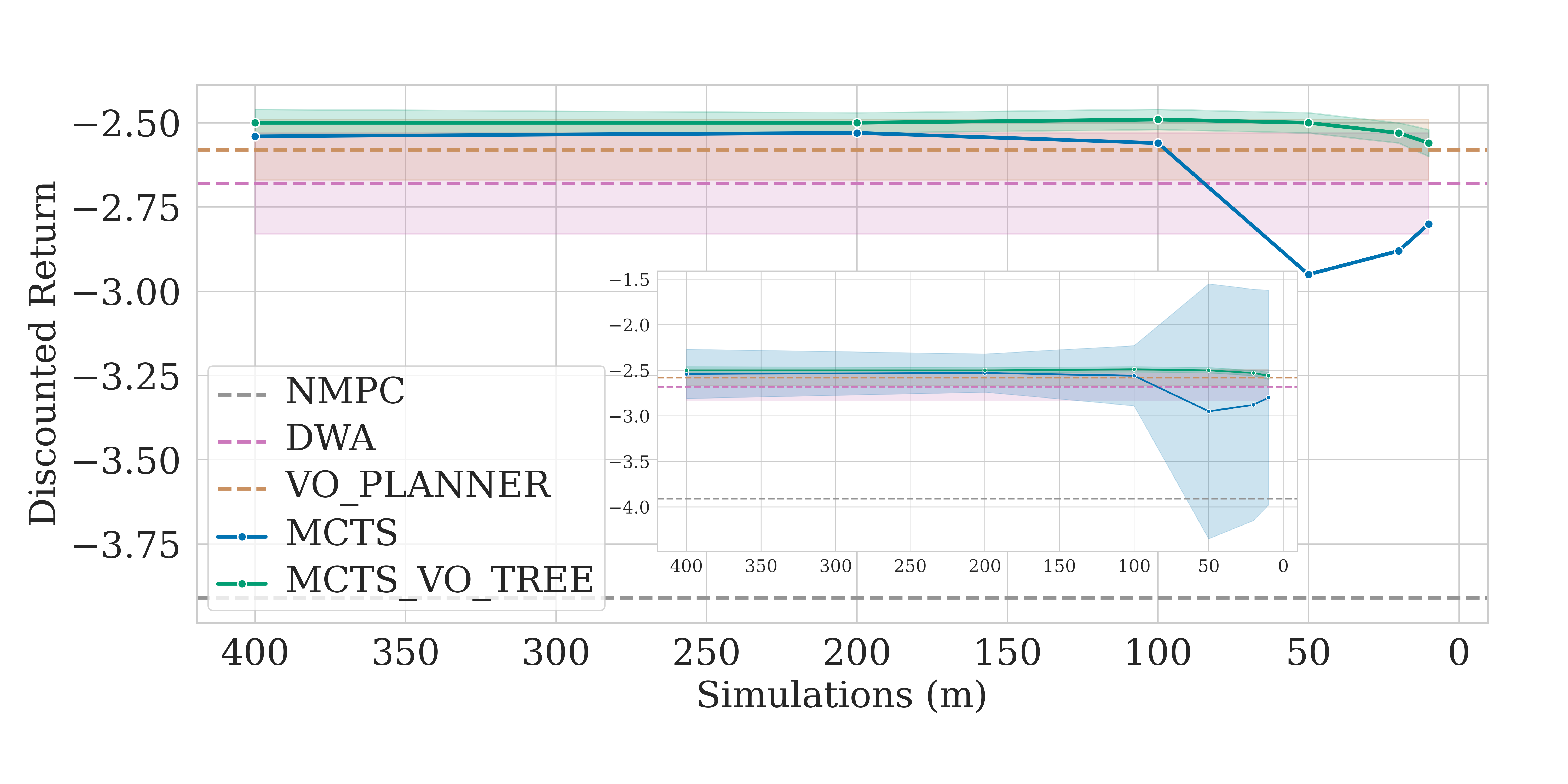}\label{fig:1confronto_rew}}
    \subfloat[Planning Time per Step ($t_{plan}$, mean $\pm$ std. dev.)]{\includegraphics[width=0.33\textwidth]{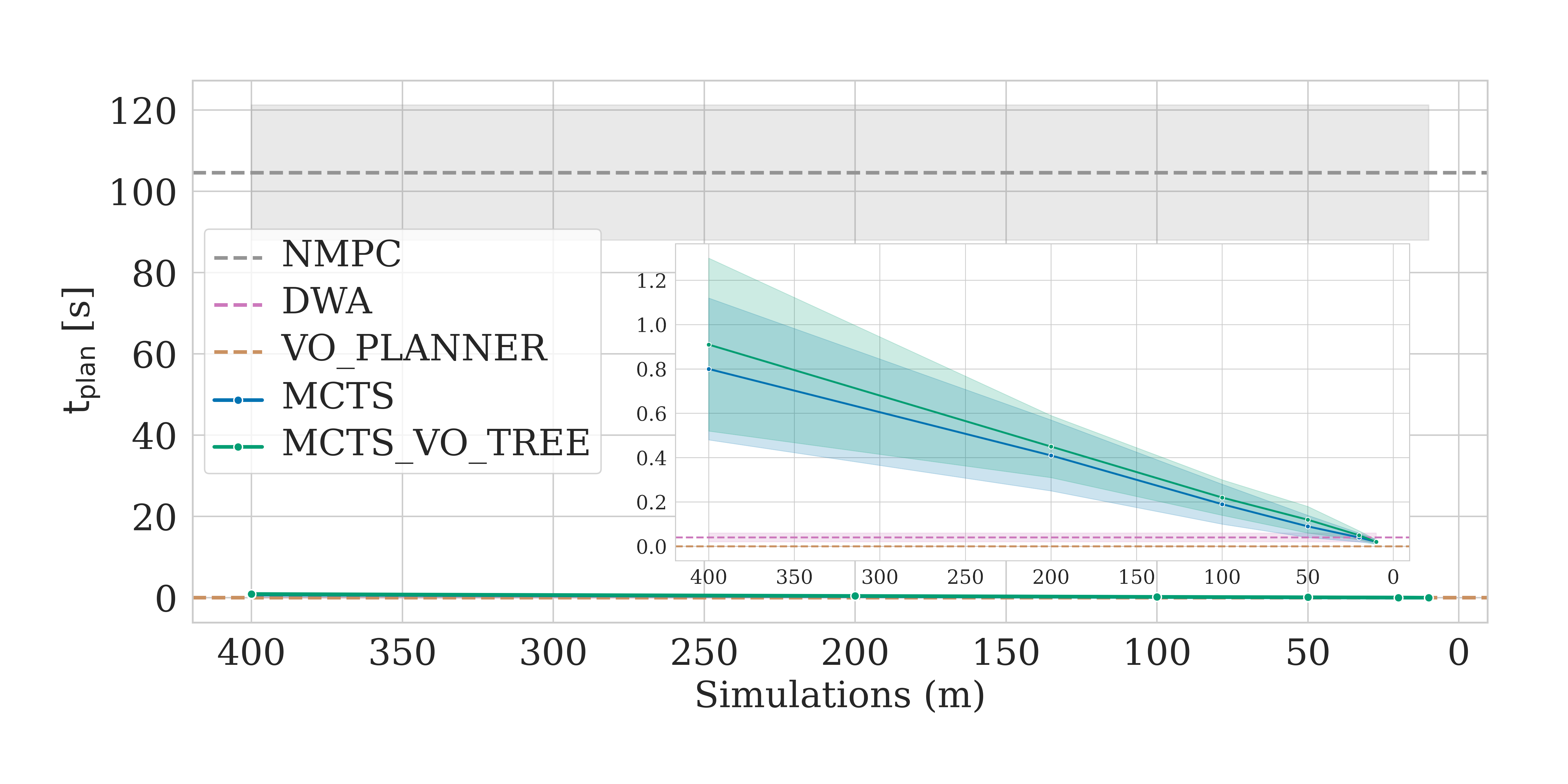}\label{fig:1confronto_time}}\\
    \caption{Comparison between MCTS\_VO\_TREE, MCTS, NMPC, DWA and VO\_PLANNER ($m$ is the number of MCTS simulations)}\label{fig:confronto}
\end{figure*}
We compare the performance of MCTS\_VO\_TREE (i.e., the best version among the proposed approaches) with that of NMPC, DWA\footnote{We use the available Python implementations at \href{https://github.com/atb033/multi_agent_path_planning}{https://github.com/atb033/multi\_agent\_path\_planning} (NMPC) \cite{liu2017path} and \href{https://github.com/AtsushiSakai/PythonRobotics/}{https://github.com/AtsushiSakai/PythonRobotics/} (DWA)}, VO\_PLANNER and MCTS.
Example GIFs showing the performance of different algorithms are available in the supplementary material.
Curves for NMPC (grey lines), DWA (pink lines) and VO\_PLANNER (yellow lines) do not change with the number of simulations, since they are independent on that. Hyperparameter tuning was conducted for NMPC, in order to find the best time horizon $\tau \in [10, 70]$ to balance between computational efficiency and task performance, resulting in $\tau = 70$.

Figure \ref{fig:1confronto_rew}\footnote{For NMPC, we omit the standard deviation since it is very large and affects readability. We include the plot with the standard deviation in the supplementary material.} shows that MCTS\_VO\_TREE (green curve) outperforms all competitors in terms of discounted return ($\rho$), especially achieving a much narrower standard deviation, i.e., more robust behaviour across 50 random trials per $m$ value. The performance remains superior also when very few simulations per step are made ($m < 20$) because VO focuses the search performed by MCTS on the collision-free action subspace avoiding useless simulations. Moreover, together with VO\_PLANNER (which however achieves lower average return), MCTS\_VO\_TREE  is the only algorithm ensuring safe collision avoidance for any number of simulations, under the assumptions in Section \ref{sec:ass}, as evidenced by Figure \ref{fig:1confronto_coll}.
In particular, the low return achieved by NMPC is probably due to the highly cluttered environments and the high rate of collisions ($> 80\%$ from Figure \ref{fig:1confronto_coll}).
Moreover, the MCTS collision-rate increases as the number of simulations decreases, evidencing the need for safe VO action pruning, especially in highly sub-optimal conditions (low value of $m$).

It is interesting to analyze the success rate of the algorithms, namely, the percentage of experiments where the goal is reached within the maximum number of steps (100) without collisions (plots are reported in the supplementary material).
Specifically, MCTS\_VO\_TREE reaches the goal in about $80\%$ of the experiments, without significant variations as $m$ decreases. VO\_PLANNER is the second best-performing planner ($70\%$), while other algorithms perform significantly worse, especially MCTS, which experiences a very low success rate of around 20\% as $m$ decreases. This proves the fundamental role of VO action pruning at efficiently guiding the agent towards the best and safest trajectories to the goal, but also the advantage of using a look-ahead planner instead of a reactive one as VO\_PLANNER.

Considering the computational performance (planning time $t_{plan}$), in Section \ref{sec:ts_condition} we mentioned that $t_{plan} < t_s = \SI{1}{s}$ in order to guarantee collision avoidance. From Figure \ref{fig:1confronto_time}, MCTS\_VO\_TREE and MCTS achieve this when simulations $m<300$, while DWA and VO\_PLANNER are the most computationally efficient. However, as explained above, MCTS\_VO\_TREE significantly otuperforms the other algorithms in terms of collision rate, discounted return and success rate. 
In addition, from Figure \ref{fig:1confronto_rew}, MCTS\_VO\_TREE achieves high return even with very few simulations ($m<20$), while MCTS performance significantly drop for $m<100$. Hence, in practice MCTS\_VO\_TREE can achieve much better computational performance with respect to MCTS, working with fewer simulations (e.g., on average $t_{plan} = \SI{0.2}{s}$ with MCTS when $m=100$, while $t_{plan} < \SI{0.1}{s}$ with MCTS\_VO\_TREE when $m<50$).  
On the other hand, NMPC exhibits a substantially longer step time of \SI{15.5}{s}, which makes it unusable with $t_s= \SI{1}{s}$. 


\subsection{Ablation study}\label{sec:ablation}
\begin{figure*}
    \centering
    \subfloat[Collision Rate ($\eta$)]{\includegraphics[width=0.33\textwidth]{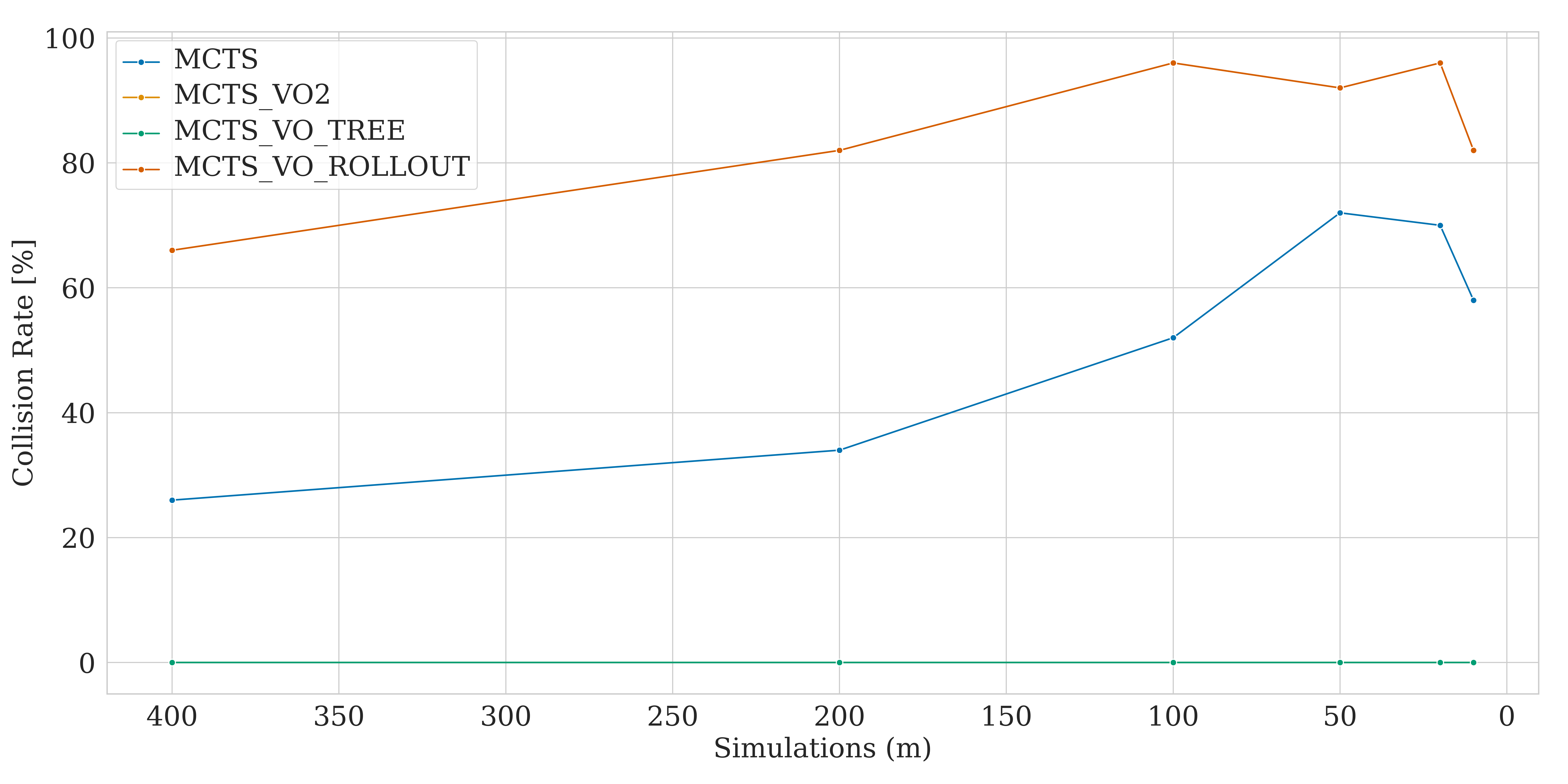}\label{fig:1ablation_nocoll}}
    \subfloat[Discounted Return ($\rho$)]{\includegraphics[width=0.33\textwidth]{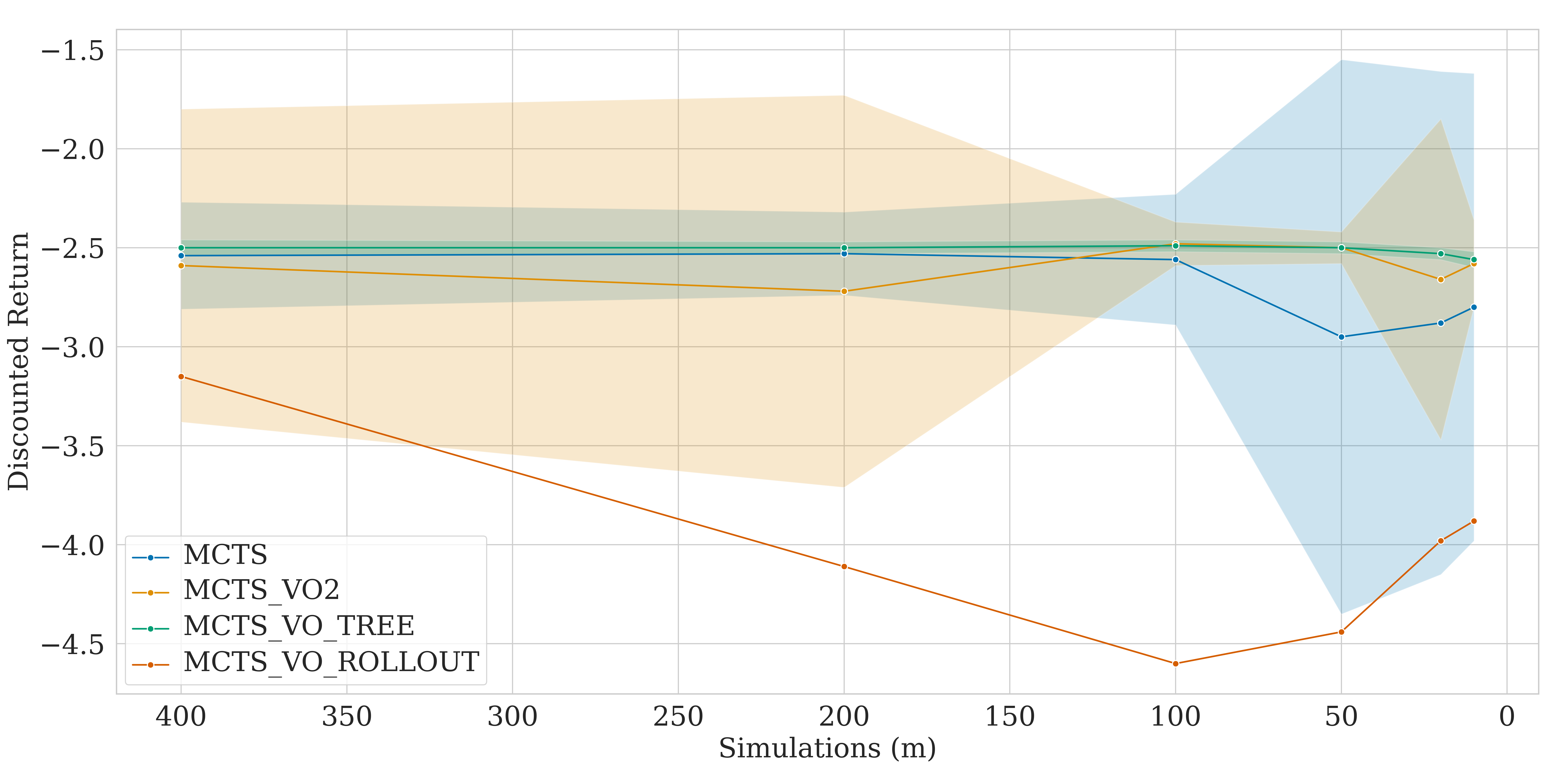}\label{fig:1ablation_rew}}
    \subfloat[Planning Time per Step ($t_{plan}$)]{\includegraphics[width=0.33\textwidth]{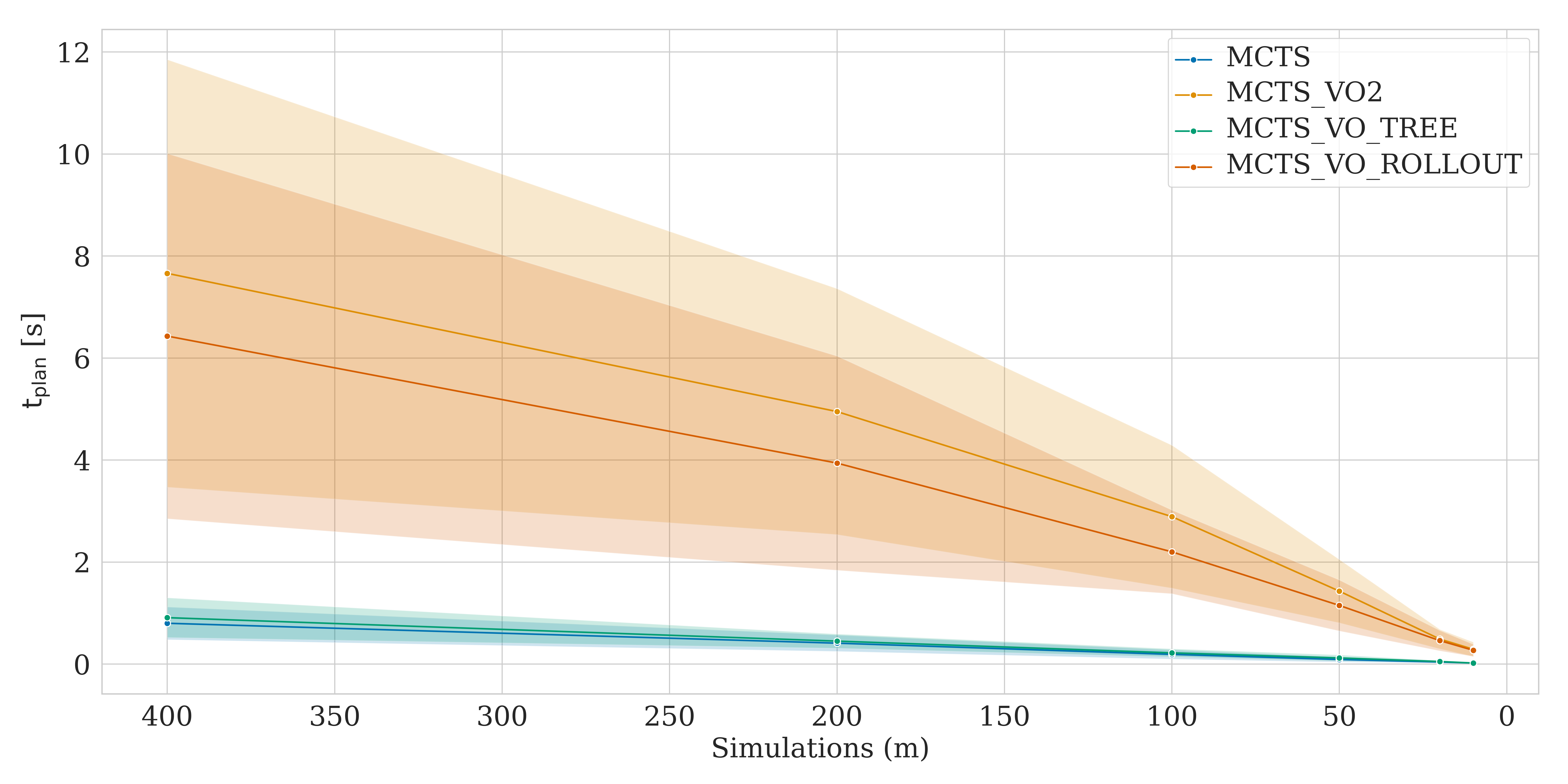}\label{fig:1ablation_time}}\\
    \caption{\label{fig:ablation}Results of the ablation study and comparison between MCTS\_VO\_TREE, MCTS, MCTS\_VO2 and MCTS\_VO\_ROLLOUT.}
\end{figure*}

To better understand the effects of the introduction of VO in different points of the MCTS algorithm, we compare the performance of four variants of the proposed algorithm, namely, MCTS\_VO\_TREE, MCTS\_VO\_ROLLOUT, MCTS\_VO2, and MCTS.
The results are depicted in Figure \ref{fig:ablation}. Each line represents a different approach: blue for MCTS, orange for MCTS\_VO2, red for MCTS\_VO\_ROLLOUT, and green for MCTS\_VO\_TREE. 
Figure \ref{fig:1ablation_rew} shows that MCTS\_VO\_TREE achieves the best discounted return on average, with the smallest standard deviation, even with very few simulations ($m < 20$), thus confirming the results in Figure \ref{fig:1confronto_rew}.
MCTS\_VO2 has similar performance on average, but with higher standard deviation, hence being less stable. This is probably due to the lower success rate ($< 70\%$, vs. $80\%$ achieved by MCTS\_VO\_TREE). Indeed, VO in the rollout phase (Algorithm \ref{alg:vor}) limits the exploration of the action space, thus the agent more often chooses to remain stationary (null velocity) since no actions are available deep in the tree (Line 16 in Algorithm \ref{alg:cc}).
MCTS\_VO\_ROLLOUT achieves the lowest discounted return, since it does not safely select velocities in UCT and the rollout exploration is limited by VO action pruning.
Figure \ref{fig:1ablation_nocoll} shows that both MCTS\_VO\_TREE an MCTS\_VO2 guarantee no collisions, even with $m=10$ simulations. Indeed, in both algorithms actions are selected according the VO constraint in UCT. On the other hand, MCTS\_VO\_ROLLOUT and MCTS do not guarantee safe collision avoidance, and their performance downgrades with fewer simulations. 

In Figure \ref{fig:1ablation_time}, the average planning time per step ($t_{plan}$) is illustrated. It is evident that integrating VO into the rollout phase (i.e., MCTS\_VO\_ROLLOUT and MCTS\_VO2) significantly increases the computational time, primarily attributed to the computational cost incurred by the computation of $\pazocal{V}_c$ (Algorithm \ref{alg:cc}) at each simulation step.
On the other hand, MCTS\_VO\_TREE only slightly increases the planning time step with respect to MCTS because it does not use VO in the computationally-expensive rollout phase.
However, the computational time increase is absorbed by the advantages in terms of planning performance (discounted return and collision rate).
Moreover, for all values of $m$, MCTS\_VO\_TREE always requires a computational time per step lower than $t_s$ (i.e., $< \SI{1}{s}$, which is a fundamental assumption to guarantee safe collision avoidance (Section \ref{sec:ts_condition}).
On the contrary, MCTS\_VO2 only meets this requirement with a very low number of simulations ($m<50$), where the discounted return is significantly less stable with higher standard deviation (Figure \ref{fig:1ablation_rew}).

Finally, we remark that the computational efficiency is particularly important for practical deployment on real robots since the number of simulations affects the computational requirements on board of the physical system.

\section{Conclusion and future works}\label{sec:conc}
We presented a novel algorithm for optimal online motion planning with collision avoidance in unknown dynamic and cluttered environments, combining the benefits of a look-ahead planner as Monte Carlo Tree Search (MCTS), with the safety guarantees about collision avoidance provided by Velocity Obstacles (VO). 
As evidenced by our ablation study, VO in the UCT phase of MCTS allows to significantly reduce the computational cost of the planner, restricting the action space to only collision-free actions and dramatically reducing the number of required online simulations (even with a large action space consisting of $\approx 60$ velocities). 
Moreover, we thoroughly discussed the assumptions required to guarantee safe collision avoidance, showing their feasibility in most practical robotic use cases.
Notably, our algorithm does not require any prior knowledge about the trajectories of other obstacles, but only their positions at the planning time and maximum velocities.
Validation in a $10\times 10$\si{m} map with up to 40 static and randomly moving obstacles shows that our approach can compute high-quality trajectories with very few simulations per step in MCTS (less than 50), maintaining low variability in random scenarios (hence being more robust) and significantly outperforming several established competitors, including NMPC and DWA. 

In future works, we will investigate the extension of our methodology to continuous action spaces and partially observable MDPs, that present additional computational and modeling challenges but are of more practical interest in robotic domains. 



\bibliographystyle{ACM-Reference-Format} 
\bibliography{sample}


\end{document}